\documentclass{article}

\usepackage[final]{neurips_2024}

\usepackage{times}
\usepackage{soul}
\usepackage{url}
\usepackage[utf8]{inputenc} 
\usepackage[T1]{fontenc}    
\usepackage{hyperref}       
\usepackage{url}            
\usepackage{booktabs}       
\usepackage{amsfonts}       
\usepackage{nicefrac}       
\usepackage{microtype}      
\usepackage{xcolor}         
\usepackage[small]{caption}
\usepackage{graphicx}
\usepackage{amsmath}
\usepackage{subfigure}
\usepackage{amsthm}
\usepackage{amssymb,amsfonts}
\usepackage{algpseudocode}
\usepackage[linesnumbered,ruled,vlined]{algorithm2e}
\usepackage{changepage}
\usepackage{subcaption}
\usepackage{multirow}

\title{Intermediate Outputs Are More Sensitive Than You Think}

\author{%
\AND{Tao Huang$^{1}$ \quad Qingyu Huang$^{1}$ \quad Jiayang Meng$^{2}$} \\
$^1$ School of Computer and Big Data, Minjiang University\\
$^2$ School of Information, Renmin University of China\\
\texttt{\{huang-tao\}@mju.edu.cn}\\
\texttt{\{3222701130\}@stu.mju.edu.cn}\\
\texttt{\{jiayangmeng\}@ruc.edu.cn}\\
}

\begin{document}

\maketitle

\begin{abstract}
  The increasing reliance on deep computer vision models that process sensitive data has raised significant privacy concerns, particularly regarding the exposure of intermediate results in hidden layers. While traditional privacy risk assessment techniques focus on protecting overall model outputs, they often overlook vulnerabilities within these intermediate representations. Current privacy risk assessment techniques typically rely on specific attack simulations to assess risk, which can be computationally expensive and incomplete. This paper introduces a novel approach to measuring privacy risks in deep computer vision models based on the Degrees of Freedom (DoF) and sensitivity of intermediate outputs, without requiring adversarial attack simulations. We propose a framework that leverages DoF to evaluate the amount of information retained in each layer and combines this with the rank of the Jacobian matrix to assess sensitivity to input variations. This dual analysis enables systematic measurement of privacy risks at various model layers. Our experimental validation on real-world datasets demonstrates the effectiveness of this approach in providing deeper insights into privacy risks associated with intermediate representations.
\end{abstract}

\section{Introduction}

Data-driven technologies have seen rapid growth, leading to increased attention on the privacy implications of deep computer vision models\cite{voulodimos2018deep, chai2021deep, brunetti2018computer, gopalakrishnan2017deep} that often process large, sensitive datasets. Although these models excel at extracting meaningful patterns from visual data, they may inadvertently expose sensitive information from the underlying training data, thereby posing significant privacy risks\cite{yeom2018privacy, orekondy2017towards, papernot2018sok}. This issue becomes particularly relevant in scenarios involving gradient inversion\cite{huang2021evaluating, jeon2021gradient, hatamizadeh2022gradvit, zhang2022survey} or membership inference attacks\cite{hu2022membership, shokri2017membership, carlini2022membership, truex2019demystifying, ye2022enhanced}, where adversaries aim to reconstruct or infer sensitive data from the model’s intermediate outputs.

A substantial body of research has proposed various methods for measuring privacy risks and enhancing data protection, such as differential privacy\cite{dwork2006differential, dwork2008differential, abadi2016deep, wei2020federated}, k-anonymity\cite{mahanan2021data, saxena2022enhancing, slijepvcevic2021k}, and l-diversity\cite{parameshwarappa2021anonymization, ashkouti2021di, mehta2022improved, gangarde2021privacy}. However, these approaches primarily focus on the overall model outputs and often overlook the privacy vulnerabilities within the intermediate layers of deep learning models. Intermediate representations are essential for capturing and processing information, yet they may retain substantial details about the input data, increasing the risk of privacy leakage\cite{sun2021soteria, mireshghallah2020privacy}.

Many privacy assessment methods rely on attack simulations\cite{ucedavelez2015risk, johnson2018meta}, such as model inversion, gradient inversion, and membership inference attacks (MIA), to evaluate whether an adversary could exploit intermediate results. However, these attack-based methods face significant limitations. First, they require additional attack simulations, which are often computationally expensive, particularly in large-scale computer vision systems. Second, these methods are inherently incomplete; exhaustively testing for all potential attacks is impractical, limiting their reliability for comprehensive privacy risk assessment.

To address these limitations, there is a need for more systematic and computationally efficient privacy risk assessment methods for deep computer vision models. In this paper, we propose a method for classifying the sensitivity levels of intermediate outputs without relying on adversarial simulations. Our approach assesses sensitivity levels during the model’s training phase, allowing model developers to identify and evaluate the sensitivity of intermediate results in real-time, thereby enabling systematic privacy risk assessment.

We introduce the concept of Degrees of Freedom (DoF)\cite{campbell1975iii, toraldo1969degrees, pal2010nested}, commonly used in statistical modeling to quantify model complexity, as a novel metric for privacy risk classification. Layers with lower DoF may retain more specific details about the input data, potentially posing a greater privacy risk. To fully capture the privacy risks, we also analyze the sensitivity of intermediate results to input perturbations, which we quantify using the rank of the Jacobian matrix. This combined analysis provides a comprehensive framework for assessing the privacy risks of intermediate outputs without the need for attack simulations.

The primary research question we address in this study is: \textit{How can data privacy levels be classified based on the Degrees of Freedom and sensitivity of intermediate outputs in a deep learning model?} We develop a framework that leverages both DoF of intermediate layer outputs and the Jacobian rank with respect to input data to classify privacy sensitivity across layers. This approach offers a deeper understanding of the risks associated with intermediate layers, often neglected in conventional privacy assessment methodologies, and serves as an efficient alternative to attack-based classifications.

Our contributions are threefold: \begin{itemize} \item \textbf{Introduction of Degrees of Freedom as a Metric for Privacy Risk Assessment:} We propose using the Degrees of Freedom (DoF) of intermediate layer outputs as a novel means to quantify information retention about input data, thereby offering a new perspective on privacy leakage risks at various stages in deep computer vision models. \item \textbf{Development of a Privacy Classification Framework Based on DoF and Jacobian Rank:} We present a privacy classification framework that combines DoF analysis with the sensitivity of intermediate outputs, measured via Jacobian rank. This dual approach provides a more detailed and accurate privacy risk assessment than current methods. \item \textbf{Experimental Validation on Real-World Datasets:} We validate the effectiveness of our framework through empirical testing on real-world datasets. Our results illustrate how the combination of DoF and Jacobian rank offers deeper insights into privacy risks associated with intermediate layers in deep computer vision models. \end{itemize}

\section{Related Work}
\subsection{Overview of Data Privacy Metrics}
A significant body of research has focused on developing privacy-measuring and privacy-preserving techniques to mitigate the risks of sensitive data exposure in deep learning models. Differential privacy (DP)\cite{dwork2006differential, dwork2008differential, abadi2016deep, wei2020federated} is one of the most widely adopted frameworks, ensuring that individual records within a dataset do not significantly affect the model's output. The primary strength of differential privacy is its mathematical rigor. DP provides formal guarantees about the privacy of individual data points. K-anonymity\cite{mahanan2021data, saxena2022enhancing, slijepvcevic2021k} ensures that an individual’s data cannot be distinguished from at least k-1 other individuals in the dataset, while l-diversity\cite{parameshwarappa2021anonymization, ashkouti2021di, mehta2022improved, gangarde2021privacy} ensures that sensitive information within these groups maintains diversity. These methods are relatively simple to implement and can be effective in structured datasets. However, they are often less applicable to unstructured data commonly used in computer vision tasks and may be susceptible to attacks such as attribute inference or background knowledge attacks.

While these methods offer useful privacy measurements, they primarily focus on the final outputs of a model or the data used during training. They do not address the potential leakage measurements of sensitive information through intermediate results, which are often ignored in these privacy frameworks. This gap necessitates the development of privacy metrics that can evaluate the sensitivity of intermediate representations, as addressed in this paper.

\subsection{Degrees of Freedom in Statistical and Machine Learning Models}

The concept of Degrees of Freedom (DoF)\cite{campbell1975iii, toraldo1969degrees, pal2010nested} is well-established in statistical modeling, where it refers to the number of independent parameters that a model can adjust to fit data. In machine learning, DoF is related to the model’s complexity and flexibility: a model with fewer degrees of freedom can capture more detailed patterns in the data, but it may also risk overfitting and memorizing specific training examples, which could lead to privacy leakage. This relationship between overfitting suggests that models with lower DoF may retain more specific and detailed information about the input data, posing greater privacy risks.

In the context of deep learning, research on DoF is relatively nascent. Recent studies have explored the use of DoF to assess model generalization\cite{zhang2021understanding, zhou2022domain, wang2022generalizing}, showing that models with lower DoF tend to perform better on specific tasks. However, there has been limited into the application of DoF for privacy analysis, particularly for classifying the sensitivity of intermediate outputs. This study builds on the idea that lower DoF in intermediate layers may correlate with greater privacy risks, proposing a novel approach for privacy classification based on DoF.

\subsection{Privacy Classification Frameworks}
Existing privacy classification frameworks often rely on attack-based evaluations to determine the vulnerability of models to privacy breaches. These methods typically involve simulating specific attacks, such as model inversion or membership inference, to assess how much information an adversary can extract from the model's intermediate or final outputs. For example, \cite{shokri2017membership} proposed a framework for membership inference attacks, demonstrating that models can leak information about whether specific data points were part of the training set. While such attack-based methods provide insights into model vulnerability, they are computationally expensive and inherently limited in scope. It is impossible to account for every potential attack scenario, which leaves these frameworks incomplete for evaluating the full range of privacy risks.

Alternative approaches to privacy classification have focused on measuring the sensitivity of model gradients\cite{borgonovo2016sensitivity, ancona2017towards} or the impact of input perturbations\cite{papernot2016distillation, ivanovs2021perturbation} on model outputs, using metrics such as input-output Jacobians\cite{srinivas2018knowledge, hoffman2019robust}. These methods provide a more systematic way to evaluate privacy risks by focusing on how changes in input affect model outputs, without relying on specific attack simulations. However, while these frameworks capture certain aspects of privacy sensitivity, they do not fully account for the complexities of intermediate representations in deep learning models.

The proposed framework in this study builds on combining DoF analysis with the rank of the Jacobian matrix, providing a more comprehensive evaluation of privacy risks without the need for exhaustive attack simulations. This approach not only assesses how much information a layer retains (via DoF) but also evaluates the sensitivity of intermediate outputs to input perturbations (via Jacobian rank), offering a dual perspective on privacy classification.

\subsection{Gap Analysis}
Despite the significant advances in privacy assessment techniques, there remain critical gaps in the literature regarding the classification of privacy risks for intermediate results in deep learning models. Current privacy frameworks, whether based on differential privacy or attack simulations, are often focused on either the final output of a model or the raw data used during training. They do not provide sufficient attention to how intermediate representations, which are central to the internal workings of deep learning models, can reveal sensitive information.

Additionally, existing methods for privacy assessment that rely on attack simulations are computationally intensive and incomplete. These methods require running multiple attack tests, which incur high costs and may still fail to capture the full range of privacy vulnerabilities present in real-world applications. There is a need for efficient and systematic methods that can classify privacy risks without relying on exhaustive attack simulations.
This paper addresses these gaps by proposing a novel framework for classifying the privacy sensitivity of intermediate outputs based on Degrees of Freedom and Jacobian rank. This dual analysis provides a more comprehensive and computationally efficient approach to privacy classification, offering insights that traditional privacy metrics and attack-based frameworks cannot provide.

\section{Methodology}
In this section, we present the overall research strategy used to classify the privacy sensitivity of intermediate representations in deep learning models, focusing on the DoF and Jacobian matrix rank estimation. The study aims to evaluate how much information each layer retains about the training data and how sensitive the model's intermediate outputs are to variations in the input. 

\subsection{Why Intermediate Results Are Sensitive}
In deep computer vision models, intermediate layer outputs, also known as hidden representations, are transformations of the input data. These representations may retain detailed information about the original input, especially in over-parameterized models. Since these layers are neither explicitly optimized for privacy nor designed to mask sensitive information, they can inadvertently reveal significant details about the training data, making them privacy-sensitive.

Analyzing the Degrees of Freedom (DoF) of these intermediate representations provides a way to quantify how much of the input data’s variability is captured in each layer. In statistical terms, DoF represents the effective number of independent parameters used by a layer to describe the data. High DoF in a layer suggests that the model is capturing more specific details of the input data. While DoF can indicate the complexity of intermediate representations, it alone cannot fully capture the sensitivity of these outputs to changes in the input. Privacy risks are not only associated with how much information a layer holds but also with how that information responds to changes in the input. To address this gap, we introduce the rank of the Jacobian matrix, which measures the sensitivity of the intermediate outputs to input perturbations. A higher Jacobian rank implies that small changes in the input can lead to large variations in the output, suggesting that the intermediate representations are more vulnerable to input-based privacy attacks. Therefore, analyzing both the DoF and the Jacobian rank provides a more comprehensive assessment of the privacy risk associated with intermediate layers.

\begin{algorithm}
    \caption{\textbf{DoFs estimation of intermedia layers}}
    \label{alg: DoF}
    \KwIn{Model $F(\mathbf{x}; W)$ with parameters $W$, Dataset $\mathcal{X}$, Set of intermediate layers \( \{l\} \), whose Degrees of Freedom (DoF) need to be estimated, Batch size \( m \).}
    \KwOut{A series of \( \text{DoF}_{t}^{(l)} \), where \( t \) represents the \( t \)-th epoch, and \( l \) represents the \( l \)-th layer.}
    \textit{// Layer Output Calculation for Batch Data.} \\
    \For{$t = 1, 2, \dots,T$}{
         Calculate the output of the \( l \)-th layer \( h_t^{(l)} \) for batch \( \mathcal{B}_t \).\\
         Calculate the output matrix of the \( l \)-th layer: \[H^{(l)} = \left[ h_1^{(l)}, h_2^{(l)}, \dots, h_m^{(l)} \right] \in \mathbb{R}^{k_l \times m}\], where \( k_l \) is the output dimension of the \( l \)-th layer.\\
         Update the parameters \( W \).\\
         \For{$l = 1, 2, \dots,L$}{
         \textit{// Centralization and Projection.} \\
         \textit{// Calculate in Parallel.} \\
         Calculate the mean vector:
    \[
    \mu^{(l)} = \frac{1}{m} \sum_{j=1}^{m} h_j^{(l)}
    \]\\
    Centralize the intermediate matrix \( H^{(l)} \):
    \[
    \tilde{H}^{(l)} = H^{(l)} - \mu^{(l)} \mathbf{1}^T
    \]
    , where \( \mathbf{1}^T \) is a row vector of ones with length \( m \).\\
    Generate a random Gaussian matrix \( R^{(l)} \in \mathbb{R}^{k_l \times r_l} \), where \( r_l \ll k_l \).\\
    Compute the projected matrix:
    \[
    \hat{H}^{(l)} = R^{(l)T} \tilde{H}^{(l)} \in \mathbb{R}^{r_l \times m}
    \]\\
    \textit{// Covariance Matrix and Eigenvalue Decomposition.} \\
    Calculate the covariance matrix from the projected matrix \( \hat{H}^{(l)} \):
    \[
    C^{(l)} = \frac{1}{m} \hat{H}^{(l)} \hat{H}^{(l)T} \in \mathbb{R}^{r_l \times r_l}
    \]\\
    Perform the eigenvalue decomposition of the covariance matrix \( C^{(l)} \):
    \[
    \lambda_1^{(l)}, \lambda_2^{(l)}, \dots, \lambda_{r_l}^{(l)}
    \]
    , where \( \lambda_1^{(l)} \geq \lambda_2^{(l)} \geq \cdots \geq \lambda_{r_l}^{(l)} \geq 0 \) are the eigenvalues of \( C^{(l)} \).\\
    Set a threshold value \( \tau^{(l)} \).\\
    Estimate the Degrees of Freedom (DoF) for layer \( l \) at iteration \( t \):
    \[
    \text{DoF}_{t}^{(l)} = \arg \min_r \left( \sum_r \frac{\lambda_r^{(l)}}{\sum_{i=1}^{r_l} \lambda_i^{(l)}} \geq \tau^{(l)} \right)
    \]\\
    
         }
    Record the estimated Degrees of Freedom \( \text{DoF}_{t}^{(l)} \) for each layer \( l \).\\
         }
    \Return \( \{\text{DoF}_1^{(l)}, \cdots, \text{DoF}_T^{(l)} \}_{l=1}^L\).
\end{algorithm}

\subsection{Assumptions and Constraints}
The proposed framework assumes that the deep learning model has multiple layers with accessible intermediate outputs. This means that during the training process, we can retrieve and analyze the hidden layer representations for each batch of data. This is typically the case in feedforward architectures, convolutional networks, and some types of recurrent models. We also assume that the input data is representative of the underlying distribution that the model is being trained on. If the training data is heavily biased or unrepresentative, the analysis of DoF and Jacobian rank might yield misleading conclusions. The sensitivity of the intermediate outputs could be underestimated or overestimated based on the characteristics of the data distribution. The framework assumes that the data used during training is stationary, meaning its statistical properties do not change over time or across different portions of the dataset. If the data distribution shifts significantly during training (e.g., in non-stationary environments), the covariance matrix and sensitivity estimates might not accurately reflect the model’s behavior across different phases of training.

\subsection{Analysis Framework and Algorithms}

\textbf{Degrees of Freedom (DoF) Estimation.} 
Algorithm \ref{alg: DoF} offers a systematic approach to estimate the DoF of intermediate layers throughout the training of a neural network model $F(\mathbf{x}; W)$. The core idea of the algorithm is to monitor the effective dimensionality of the layer outputs as training progresses. At each training iteration, Algorithm \ref{alg: DoF} computes the outputs of specified intermediate layers for the current batch of data. 

To handle the high dimensionality of neural network outputs and reduce computational complexity, Algorithm \ref{alg: DoF} employs a random Gaussian projection. This projection maps the centralized high-dimensional data onto a lower-dimensional subspace while approximately preserving the pairwise distances between data points (a property known as the \textbf{Johnson-Lindenstrauss Lemma\cite{frankl1988johnson, matouvsek2008variants, larsen2017optimality}}). This step ensures that the essential structural information of the data is retained in a more computationally manageable form.

Once projected, Algorithm \ref{alg: DoF} calculates the covariance matrix of the lower-dimensional data. The covariance matrix encapsulates how the variables (in this case, the dimensions of the projected data) vary with respect to each other, highlighting the underlying structure of the data distribution. By performing an eigenvalue decomposition of the covariance matrix, the algorithm obtains a set of eigenvalues that represent the amount of variance captured along each principal component (direction) in the projected space.

The DoF for a layer at a given training iteration is estimated by determining the minimum number of principal components required to explain a predefined proportion $\tau^{(l)}$ (e.g., 95\%) of the total variance. This is achieved by cumulatively summing the sorted eigenvalues until the threshold $\tau^{(l)}$ is reached. The resulting DoF reflects the intrinsic dimensionality of the data representation at that layer, effectively quantifying how complex or simplified the learned features are at that point in training.

\textbf{Jacobian Rank Estimation.} The Jacobian matrix $J^{(l)}$ reflects the effective dimensionality of the layer's mapping and indicates how many independent directions in the input space significantly influence the outputs. Algorithm \ref{alg: Rank} provides a systematic approach to estimate the rank of the Jacobian matrices of intermediate layers within a neural network model $F(\mathbf{x}; W)$. Instead of computing the full Jacobian—which is often computationally infeasible for high-dimensional data—Algorithm \ref{alg: Rank} employs a combination of random projections and automatic differentiation to approximate the rank efficiently.

For each intermediate layer $l$, Algorithm \ref{alg: Rank} generates a set of Gaussian random vectors $\{ v_j^{(l)} \in \mathbb{R}^{k_l} \}_{j=1}^k$. These vectors are used to project the high-dimensional output \( h^{(l)} \) of the layer onto a lower-dimensional subspace. This projection simplifies the analysis while retaining essential information about the output variations. Algorithm \ref{alg: Rank} conducts gradient computation via automatic differentiation. It calculates the inner products \( s_{j}^{(l)} = \langle h^{(l)},v_j^{(l)} \rangle \) and then calculates their gradients with respect to the input $\mathbf{x}$, yielding gradient vectors. By assembling the gradient vectors into a matrix $U^{(l)}$ and computing the Gram matrix $G^{(l)}$, Algorithm \ref{alg: Rank} encapsulates the pairwise correlations between the gradients. The eigenvalues $\lambda_{i}^{(l)}$ represent the variance captured along each principal component. Algorithm \ref{alg: Rank} estimates the rank of the Jacobian as the smallest number of principal components required to explain a significant portion of the total variance (e.g., 95\%). 

To compute \( U^{(l)} \) in parallel, the following procedure can be applied:

\textbf{1. Reshaping \( h^{(l)} \):} The output of the \( l \)-th layer, denoted as \( h^{(l)} \), originally has the shape \([m, C, H, W]\), where \( m \) is the batch size, \( C \) represents the number of channels, and \( H \) and \( W \) are the height and width of the feature map, respectively. This output can be reshaped into a matrix of shape \([m, CHW]\) to facilitate further computations.

\textbf{2. Construction of Gaussian vectors:} A set of \( k \) Gaussian random vectors \( \{ v_j^{(l)} \in \mathbb{R}^{CHW} \}_{j=1}^k \) is constructed and arranged into a matrix of shape \([CHW, k]\).

\textbf{3. Matrix multiplication:} The reshaped matrix \( h^{(l)} \) of shape \([m, CHW]\) is multiplied by the matrix of Gaussian vectors of shape \([CHW, k]\). This matrix multiplication yields a matrix \( S \) of shape \([m, k]\), where each entry \( s_{j}^{(l)} \) represents the inner product between \( h^{(l)} \) and a corresponding Gaussian vector \( v_j^{(l)} \).

\textbf{4. Gradient computation via automatic differentiation:} For each column of the resulting matrix \( S \), automatic differentiation is applied to compute the gradient \( u_{j}^{(l)} = \nabla_{\mathbf{x}} s_{j}^{(l)} \), yielding gradients of shape \([m, CHW]\).

\textbf{5. Assembly of \( U^{(l)} \):} The computed gradients \( u_{j}^{(l)} \) for \( j = 1, \ldots, k \) are combined to form a tensor \( U^{(l)} \) of shape \([m, CHW, k]\).

\textbf{6. Summation across the batch dimension:} To obtain the final matrix \( U^{(l)} \), the gradients are summed over the batch dimension, resulting in a matrix of shape \([CHW, k]\). This step aggregates the contributions from all samples in the batch, forming each column of \( U^{(l)} \).

The matrix \( U^{(l)} \) with the described dimensions is then used to construct the Gram matrix \( G^{(l)} = (U^{(l)})^T U^{(l)} \), which is essential for the subsequent rank estimation.

\begin{algorithm}[htb]
    \caption{\textbf{Rank estimation of Jacobian}}
    \label{alg: Rank}
    \KwIn{Model $F(\mathbf{x}; W)$ with parameters $W$, Dataset $\mathcal{X}$, Set of intermediate layers \( \{l\} \), whose rank of Jocabian matrix need to be estimated, Batch size \( m \).}
    \KwOut{A series of \( \text{Rank}_{t}^{(l)} \), where \( t \) represents the \( t \)-th epoch, and \( l \) represents the \( l \)-th layer.}
    \textit{// Layer Output Calculation for Batch Data.} \\
    \For{$t = 1, 2, \dots,T$}{
         Generate $k$ Gaussian random vectors \( \{ v_j^{(l)} \in \mathbb{R}^{k_l} \}_{j=1}^k \). \\
         \textit{// Calculate in Parallel.} \\
         Calculate the output of the \( l \)-th layer \( h^{(l)} \) for batch \( \mathcal{B}_t \).\\
         Calculate the inner product: \( s_{j}^{(l)} = \langle h^{(l)},v_j^{(l)} \rangle \).\\
         Calculate by auto differentiation: \( u_{j}^{(l)} = \nabla_{\mathbf{x}} s_{j}^{(l)} \).\\
         Establish the matrix:
    \[
    U^{(l)} = [u_{1}^{(l)}, \cdots, u_{k}^{(l)}]
    \]\\
        Calculate Gram matrix: \( G^{(l)} = (U^{(l)})^T U^{(l)} \in \mathbb{R}^{k \times k} \). \\
        Perform the eigenvalue decomposition of the Gram matrix \(  G^{(l)} \):
    \[
    \lambda_1^{(l)}, \lambda_2^{(l)}, \dots, \lambda_{k}^{(l)}
    \]
    , where \( \lambda_1^{(l)} \geq \lambda_2^{(l)} \geq \cdots \geq \lambda_{k}^{(l)} \geq 0 \) are the eigenvalues of \(  G^{(l)} \).\\
    Set a threshold value \( \tau^{(l)} \).\\
        \textit{// Estimate the rank of Jocabian.} \\
        \[ \text{Rank}_t^{(l)} = \arg \min_r \left( \sum_r \frac{\lambda_{r}^{(l)}}{\sum_{q=1}^k \lambda_{q}^{(l)}} \geq \tau^{(l)} \right) \]
        Record the estimated ranks \( \text{Rank}_{t}^{(l)} \) for each layer \( l \).\\
        Update the parameters \( W \).
         }
    \Return \( \{\text{Rank}_1^{(l)}, \cdots, \text{Rank}F_T^{(l)} \}_{l=1}^L\) .
\end{algorithm}

\subsection{Cross-Validation}
After employing the proposed algorithms to analyze the degrees of freedom of intermediate results in different layers and examining the behavior patterns of the rank of the Jacobian matrix of these results with respect to the input, we further validate whether these behavioral patterns are associated with the success rate of privacy attacks. Specifically, in our experiments, we recorded the changes in the degrees of freedom and ranks of various intermediate layers. By thoroughly analyzing the patterns of these changes, we evaluated the success rates of privacy attacks under different scenarios of intermediate layer result leakage. Finally, we established a correlation between the variation patterns of degrees of freedom and ranks with the success rates of privacy attacks. This analysis demonstrates how the changes in degrees of freedom and ranks can be leveraged to classify the sensitivity levels of intermediate layer outputs.

The purpose of this cross-validation is to demonstrate that our proposed method for classifying the privacy sensitivity levels of data is compatible with existing privacy attacks. This compatibility enables model trainers to monitor these indicators during training to assess the sensitivity levels of intermediate results without requiring additional privacy attacks to infer the sensitivity of these intermediate outputs.

\section{Experiment Setup}

To validate the proposed algorithms for Degrees of Freedom (DoF) estimation and Jacobian rank estimation in intermediate layers, we conducted a series of experiments using different deep learning models and datasets. These models and datasets are the backbones of computer vision in consumer electronics. Below, we detail the experimental setup, including the network architectures, datasets, training configurations, and key parameter settings. 

All experiments are performed on a server with two Intel(R) Xeon(R) Silver 4310 CPUs @2.10GHz, and four NVIDIA 4090 GPUs. The operating system is Ubuntu 22.04 and the CUDA Toolkit version is 12.4. All computer vision experimental training procedures are implemented based on the latest versions of Pytorch and Opacus, with custom functions developed for the DoF and Jacobian rank estimations.

\subsection{Model Architectures and Datasets.}
We evaluated our methods on the following vision models:

\textbf{1-layer CNN and Fully Connected Layer}: A simple layer convolutional neural network followed by a fully connected output layer was employed as an initial baseline for the experiments. This model was tested on the MNIST \footnote{https://yann.lecun.com/exdb/mnist/} dataset, with batch size 128 and the Adam optimizer. The initial learning rate is 0.01, with a momentum of 0.9.
    
\textbf{LeNet}: The LeNet architecture was trained on the CIFAR-10 and CIFAR-100 datasets, which comprise color images with 10 and 100 classes, respectively. The optimizer is Adam. The batch sizes for CIFAR-10 \footnote{https://www.cs.toronto.edu/~kriz/cifar.html} and CIFAR-100 \footnote{https://www.cs.toronto.edu/~kriz/cifar.html} are 256 and 128, respectively. The initial learning rate is 0.05, with a momentum of 0.9.
    
\textbf{AlexNet}: The AlexNet was trained on the CIFAR-100 \footnote{https://www.cs.toronto.edu/~kriz/cifar.html} dataset. This architecture provides deeper layers and more varied intermediate outputs, enabling the validation of DoF and Jacobian rank estimation across multiple layers with richer feature extraction. The initial learning rate is 0.015, with a momentum of 0.9.

\subsection{Key Parameters for DoF and Jacobian Rank Estimation}

\begin{itemize}
    \item \textbf{Gaussian Projection Matrix Size (\( r_l \))}: For DoF estimation, the Gaussian random projection matrix \( R^{(l)} \) had dimensions \( k_l \times r_l \), where \( r_l = 0.1 \times k_l \). This size was chosen to balance computational efficiency and accuracy.
    \item \textbf{Threshold for Eigenvalue Accumulation (\( \tau^{(l)} \))}: A threshold \( \tau^{(l)} = 0.95 \) was set for both algorithms to identify the minimal number of eigenvalues that account for 95\% of the total variance. This value ensures a reliable estimation of DoF and Jacobian rank while capturing significant variance.
    \item \textbf{Number of Random Vectors (\( k \))}: For Jacobian rank estimation, \( k = 0.1 \times k_l \) random Gaussian vectors were used to form the \( U^{(l)} \) matrix, facilitating the construction of the Gram matrix \( G^{(l)} \).
\end{itemize}

\begin{figure*}[htp!]
    \centering
    \vspace{-2mm}
    \subfigure[CNN(MNIST)]{
		\includegraphics[width=0.45\textwidth]{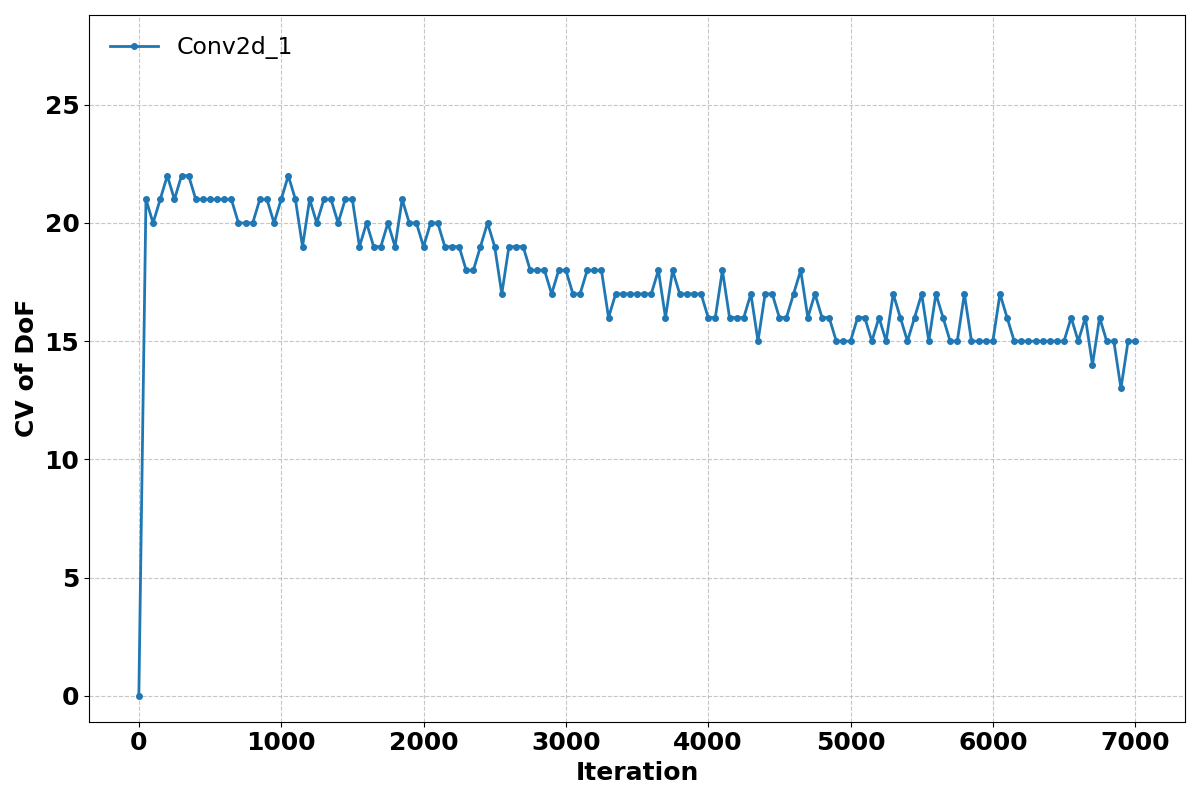}
		\label{DoF(CNN)}
    }
    \subfigure[LeNet(CIFAR10)]{
		\includegraphics[width=0.45\textwidth]{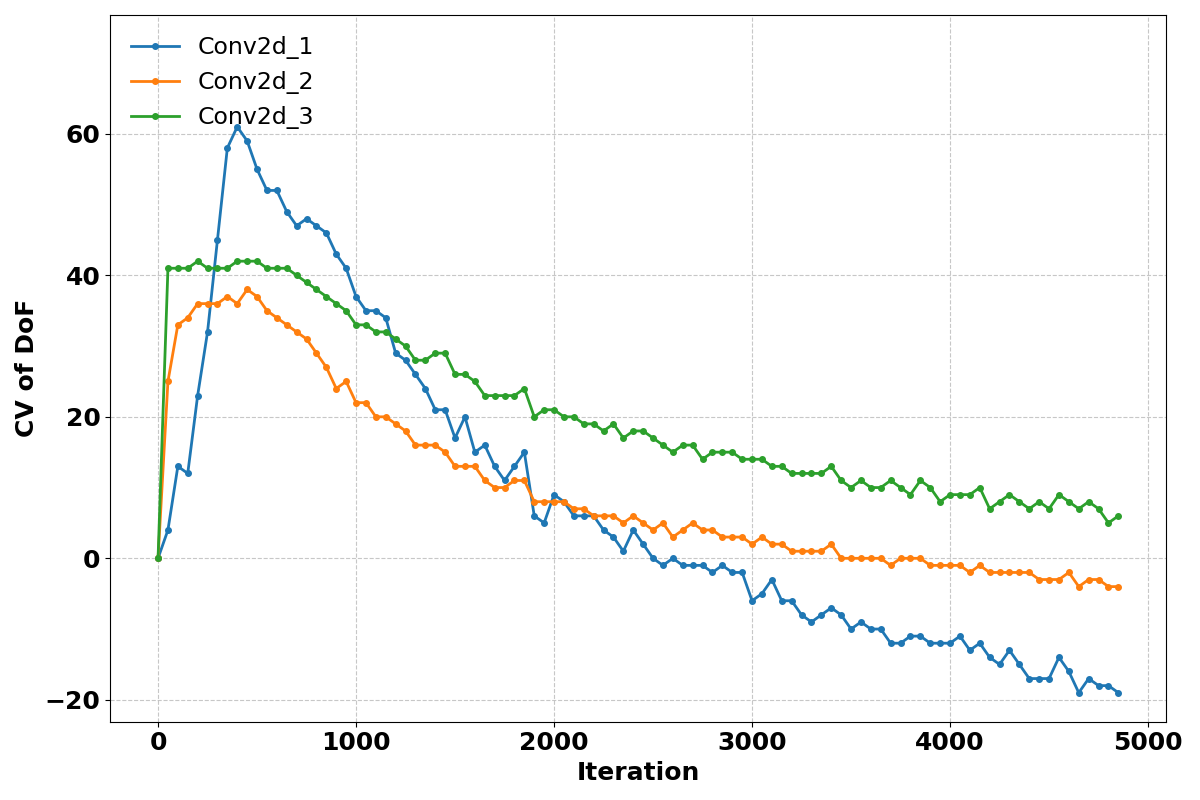}
		\label{DoF(LeNet10)}
    }\\
    \subfigure[LeNet(CIFAR100)]{
		\includegraphics[width=0.45\textwidth]{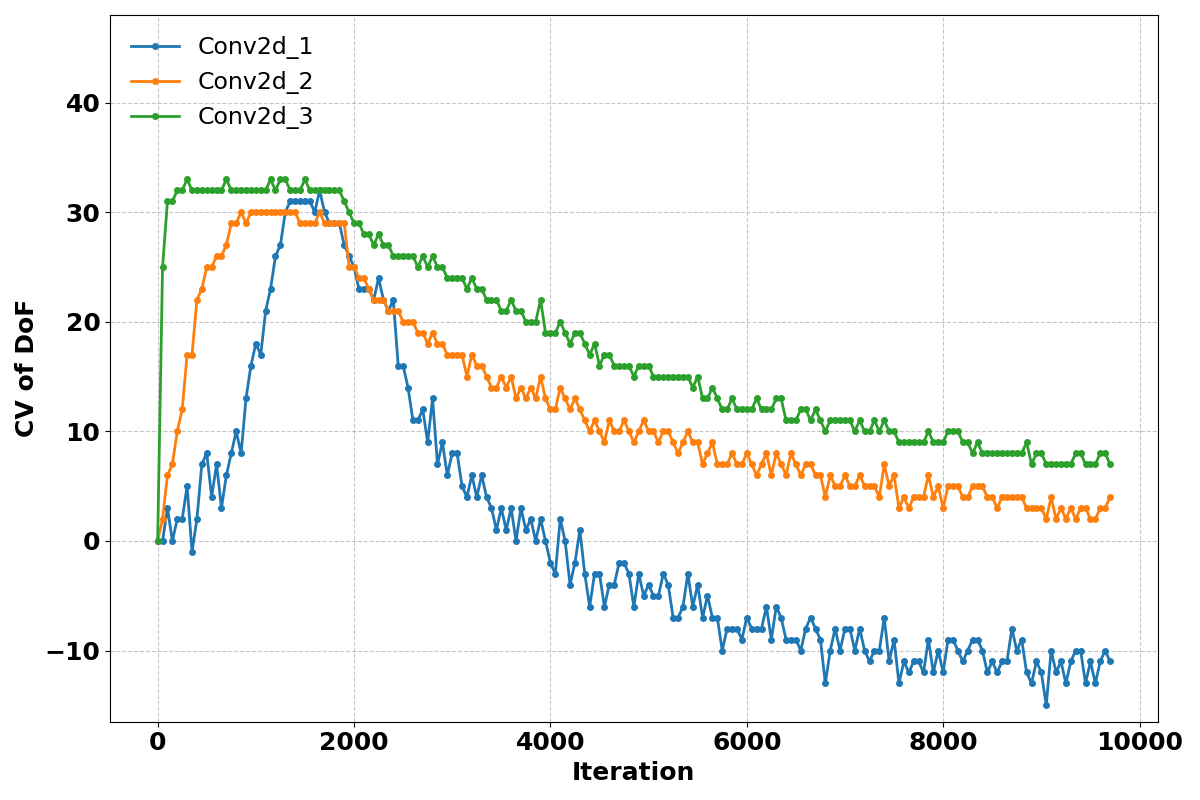}
		\label{DoF(LeNet100)}
    }
    \subfigure[AlexNet(CIFAR100)]{
		\includegraphics[width=0.45\textwidth]{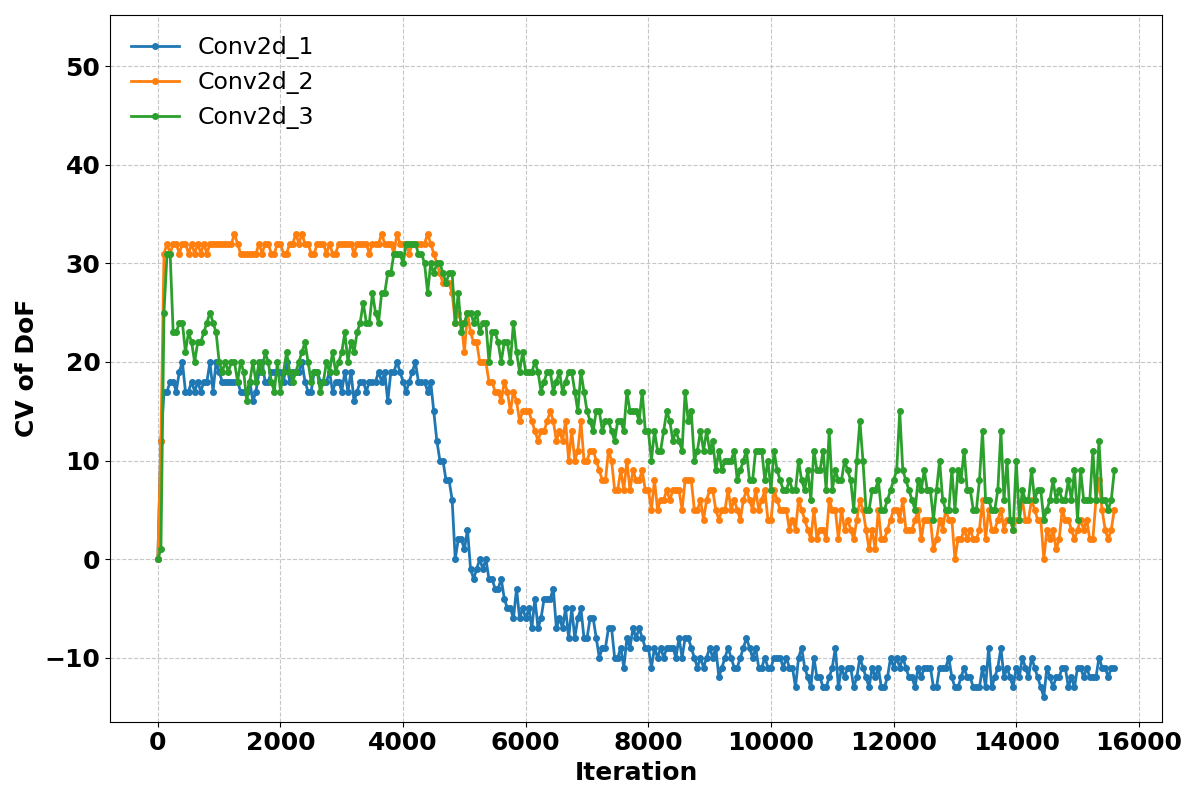}
		\label{AlexNet(LeNet1-3)}
    }
    \subfigure[AlexNet(CIFAR100)]{
		\includegraphics[width=0.45\textwidth]{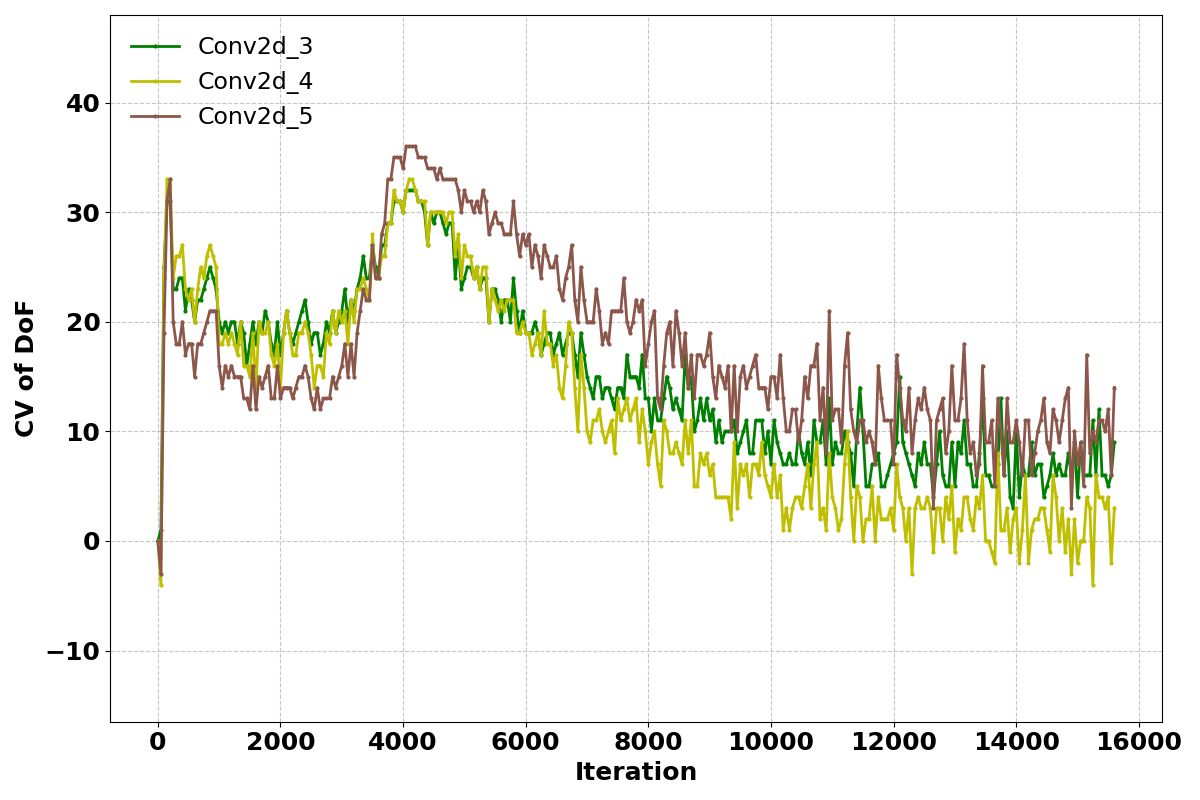}
		\label{AlexNet(LeNet3-5)}
    }
    \caption{CV of DoF}
    \label{DoF}
\end{figure*}

\begin{figure*}[htp!]
    \centering
    \vspace{-2mm}
    \subfigure[CNN(MNIST)]{
		\includegraphics[width=0.45\textwidth]{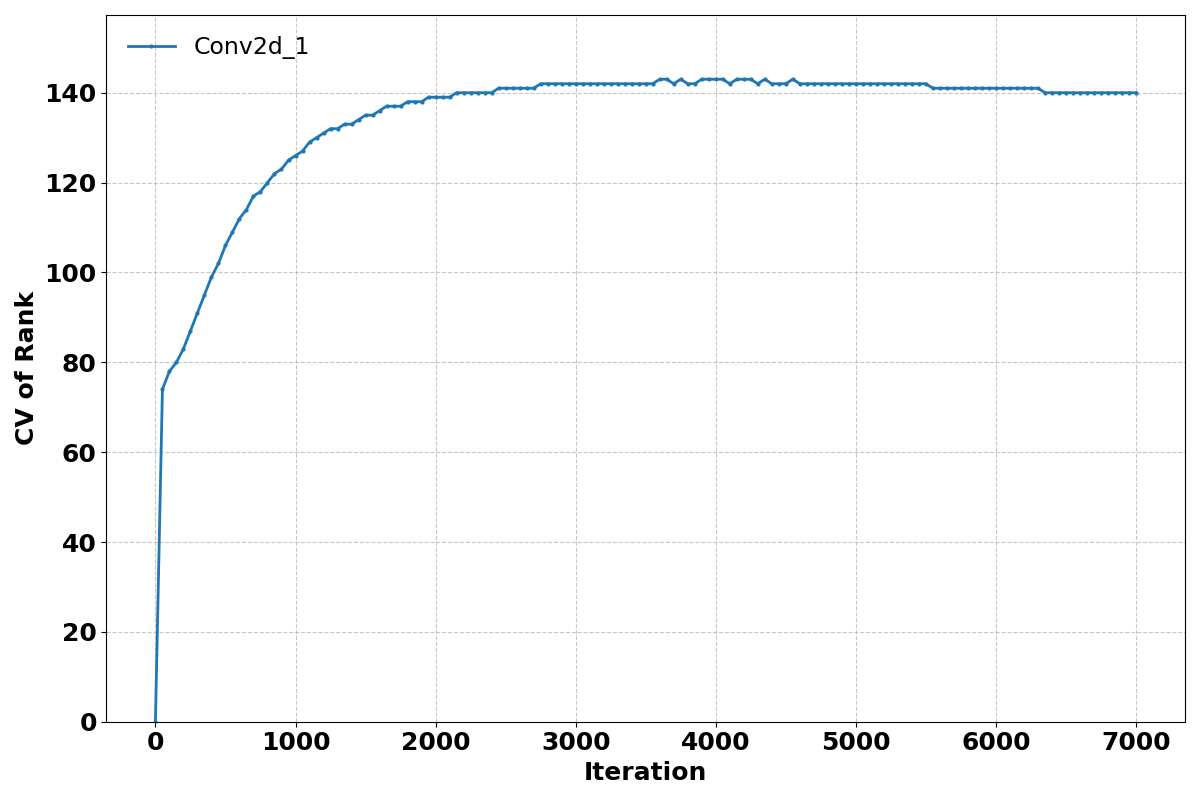}
		\label{Rank(CNN)}
    }
    \subfigure[LeNet(CIFAR10)]{
		\includegraphics[width=0.45\textwidth]{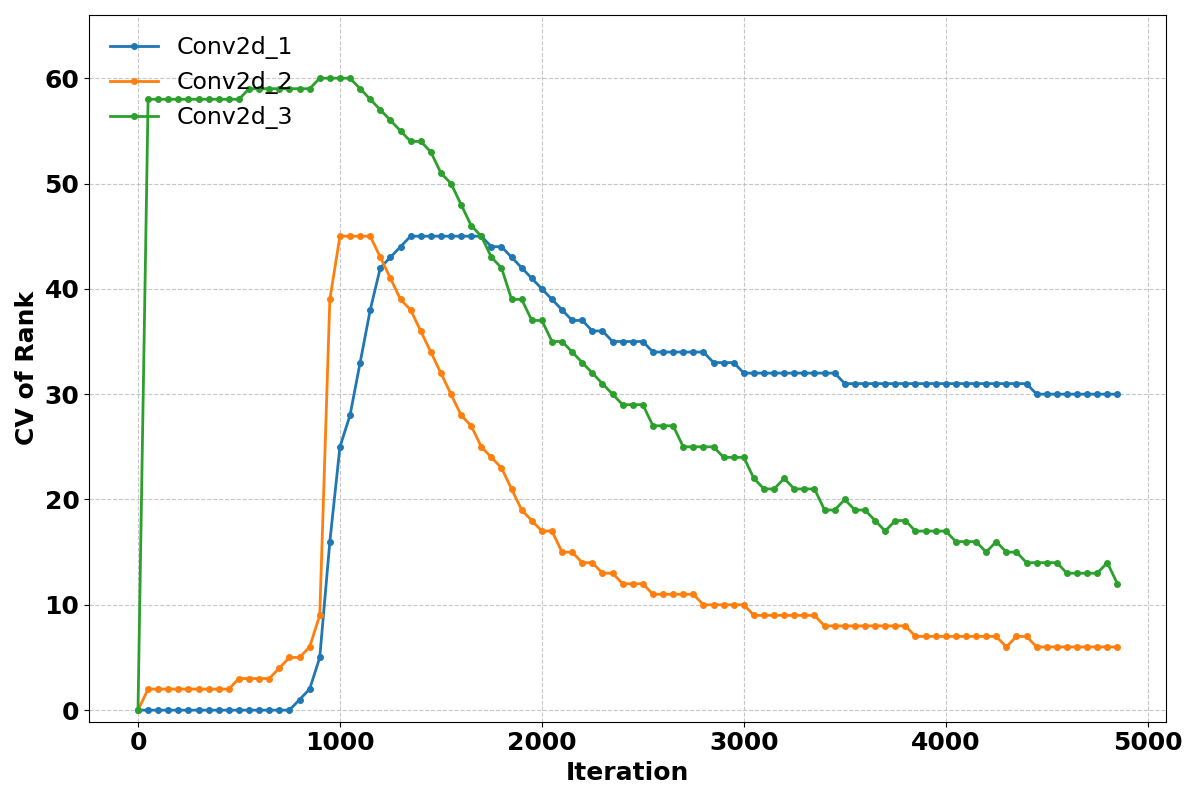}
		\label{Rank(LeNet10)}
    }\\
    \subfigure[LeNet(CIFAR100)]{
		\includegraphics[width=0.45\textwidth]{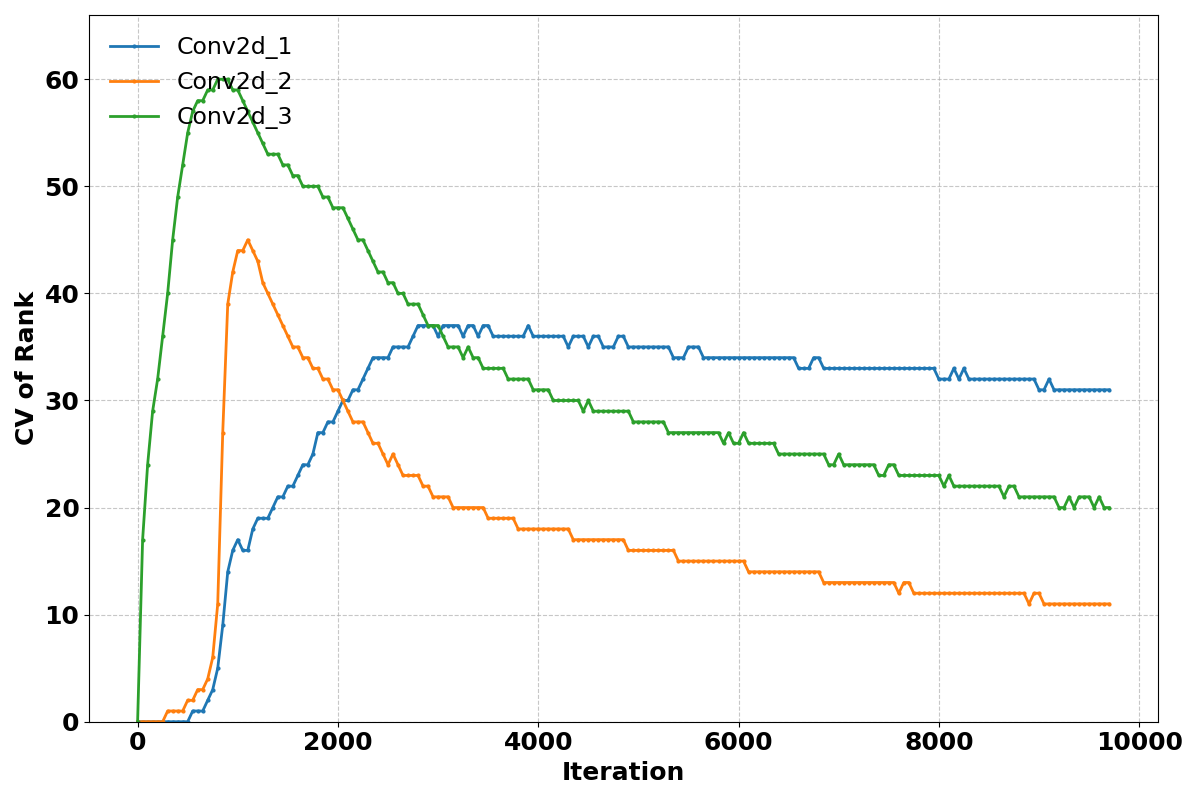}
		\label{Rank(LeNet100)}
    }
    \subfigure[AlexNet(CIFAR100)]{
		\includegraphics[width=0.45\textwidth]{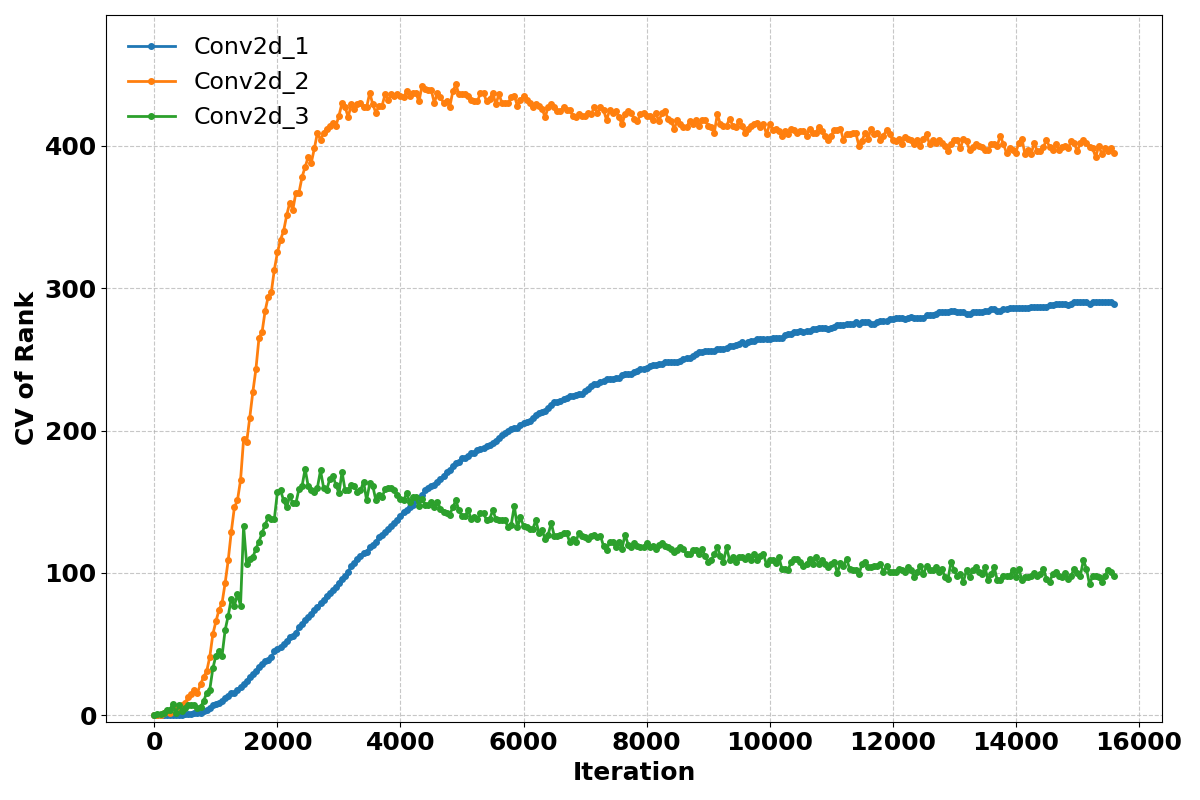}
		\label{Rank(AlexNet1-3)}
    }
    \subfigure[AlexNet(CIFAR100)]{
		\includegraphics[width=0.45\textwidth]{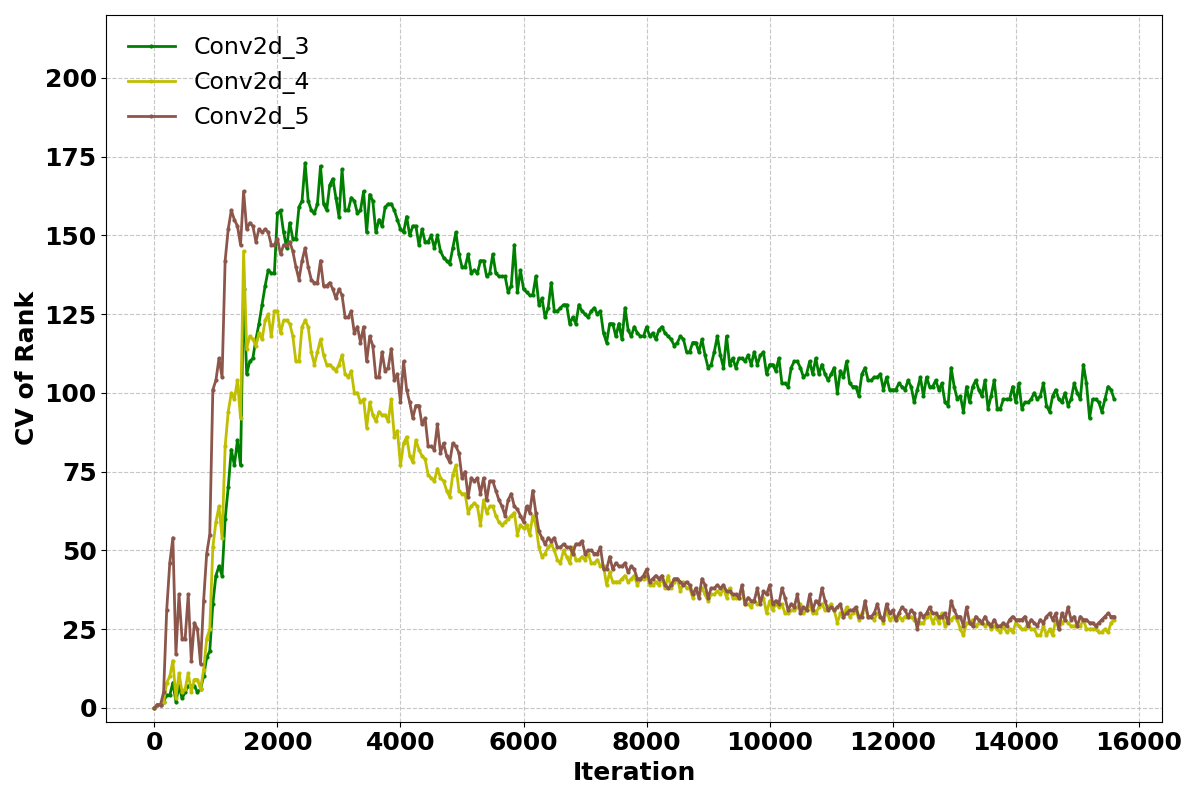}
		\label{Rank(AlexNet3-5)}
    }
    \caption{CV of Rank}
    \label{Rank}
\end{figure*}

\subsection{Membership Inference Attack Setup}

To further evaluate the relationship between DoF, Jacobian Rank, and the potential for privacy leakage, we conducted testing experiments using membership inference attacks (MIA). These attacks were designed to assess the extent to which information in the outputs of different layers contributes to the exposure of training data.

We implemented a white-box membership inference attack based on the activation outputs of intermediate layers. The attacker leveraged information from the outputs of selected layers to perform the attack, following the same setup from Nasr et al.\cite{nasr2019comprehensive} regarding layer vulnerability. We tested the outputs of individual layers to measure their impact on attack accuracy. Specifically, experiments targeted the last few layers of models, as these tend to reveal the most information about the training data.

The MIA setup is as follows. The batch size used during the attack model training was set to 64, and the Adam optimizer was employed with a learning rate of 0.0001. The attack model was trained for 100 epochs, with the best model selected based on its testing accuracy. The primary evaluation metric for the attack was the membership inference accuracy, which measures the attack model's ability to determine whether a sample was part of the training set. The attack was conducted on pre-trained models, including CNN (MNIST), LeNet (CIFAR-10, CIFAR-100), and AlexNet (CIFAR-100). For dataset partitioning, member and non-member samples were selected from both the training and test datasets, following the procedure outlined by Nasr et al. \cite{nasr2019comprehensive}, to ensure a controlled analysis of information leakage.

\begin{table*}[htbp]
\small
\begin{tabular}{c|c|c|c|c|c}
\toprule
\begin{tabular}[c]{@{}c@{}}Model \\ (Dataset)\end{tabular}               & Output Layer   & \begin{tabular}[c]{@{}c@{}}Parameter \\ Amount\end{tabular} & \begin{tabular}[c]{@{}c@{}}$\max_t \text{CV}_t^{(\text{DoF})}$ \\ ($- \text{CV}_T^{(\text{DoF})}$)\end{tabular} & \begin{tabular}[c]{@{}c@{}}$\max_t \text{CV}_t^{(\text{Rank})}$ \\ ($- \text{CV}_T^{(\text{Rank})}$)\end{tabular} & \begin{tabular}[c]{@{}c@{}}Attack \\ Accuracy\end{tabular} \\ \toprule
\multirow{1}{*}{CNN(MNIST)}        & Conv2d\_1     &                  312&24 (9) &143 (3)& 78.31\%\\
                                    \hline
\multirow{3}{*}{LeNet(CIFAR10)}    & Conv2d\_3     & 3,612&  43 (37) &60 (47)&                  72.61\%
\\
                                   & Conv2d\_2 & 3,612& 39 (43)&   46 (40)& 70.32\%                \\
                                   & Conv2d\_1  & 912&  64 (83)&   46 (16)&  69.01\%               \\ \hline
\multirow{3}{*}{LeNet(CIFAR100)}   & Conv2d\_3     & 3,612&  33 (26)&  60 (40)&                 75.16\%\\
                                   & Conv2d\_2 & 3,612&  31 (28)&  45 (34)& 73.13\%                \\
                                   & Conv2d\_1  & 912&   34 (46)&   37 (6)& 71.33\%                \\ \hline
\multirow{5}{*}{Alexnet(CIFAR100)} & Conv2d\_5     & 590,080&                         36 (28)&                                   164 (136)& 72.87\%\\
                                   & Conv2d\_4 & 884,992&                         34 (29)&                                   146 (119)& 71.05\% \\ 
                                   & Conv2d\_3  & 663,936&                         33 (25)&                                   175 (77)& 69.35\% \\ 
                                   & Conv2d\_2  & 307,292   & 33 (29)   & 447 (52)  &  68.11\%    \\
                                   & Conv2d\_1  & 23,296   &  22 (32)  & 291 (1)  &  54.76\%    \\
                                   \bottomrule
\end{tabular}
\caption{CV and Attack Accuracy}
\label{table1}
\end{table*}

\begin{table*}[htbp]
\begin{tabular}{c|c|c|c|c|c}
\toprule
\begin{tabular}[c]{@{}c@{}}Model \\ (Dataset)\end{tabular}                 & Output Layer   & \begin{tabular}[c]{@{}c@{}}Parameter \\ Amount\end{tabular} & $\text{MCR}_T^{(\text{DoF})}$ & $\text{MCR}_T^{(\text{Rank})}$ & \begin{tabular}[c]{@{}c@{}}Attack \\ Accuracy\end{tabular} \\ \toprule
\multirow{1}{*}{CNN(MNIST)}        & Conv2d\_1     &               312  &27.27\%                         &0.82\%                                   & 78.31\% \\
                                    \hline
\multirow{3}{*}{LeNet(CIFAR10)}    & Conv2d\_3     & 3,612    &  1850.00 \%                       &783.33\%                                   & 72.61\%
\\
                                   & Conv2d\_2 & 3,612   & 860.00\%                        &   181.81\%                                & 70.32\%                \\
                                   & Conv2d\_1  & 912    &  436.84\%                       &   7.14\%                                &  69.01\%               \\ \hline
\multirow{3}{*}{LeNet(CIFAR100)}   & Conv2d\_3     & 3,612   &  866.88\%                       &  666.66\%                                 &                 75.16\%\\
                                   & Conv2d\_2 & 3,612   &  466.66\%                       &  147.82\%                                 & 73.13\%                \\
                                   & Conv2d\_1  & 912  &   270.59\%                      &   2.57\%                                & 71.33\%                \\ \hline
\multirow{5}{*}{Alexnet(CIFAR100)} & Conv2d\_5     & 590,080  &                         2800.00\%&                                   1511.11\%& 72.87\%\\
                                   & Conv2d\_4 & 884,992   &2900.00\%    &                                   350.00\%& 71.05\%   \\ 
                                   & Conv2d\_3  & 663,936           & 625.00\%     &90.58\%& 69.35\%\\ 
                                   & Conv2d\_2  &  307,392  &525.00\%    &21.66\%   & 68.11\%     \\
                                   & Conv2d\_1  &  23,296  &266.60\%    & 0.10\%  &  54.76\%    \\
                                   \bottomrule
\end{tabular}
\caption{MCR and Attack Accuracy}
\label{table2}
\end{table*}

\section{Experiment results} 

\textbf{Observed trends.} In our experiments, for all models, we observed a consistent trend in the intermediate feature extraction layers where the DoF and the rank of the Jacobian matrix initially decreased and subsequently increased throughout training. Specifically, during the initial epochs, both metrics showed a marked reduction, indicating a phase of compression and abstraction in the model’s learning process. This was followed by a turning point after which the DoF and rank began to rise, suggesting an expansion in the model's ability to retain detailed input information. Although the specific epochs at which these transitions occurred varied between different models, the overall pattern was observed consistently across feature extraction layers in both models.

\textbf{Quantitative analysis.} The quantitative analysis of the DoF and Jacobian rank revealed distinct numerical changes in these metrics over training. Based on the observed trends, we utilize \textit{Change Value(CV)} and \textit{Modified Change Ratio(MCR)} of DoF and Jocabian Rank as our detailed metrics, defined in Eq.(\ref{eq1}) and Eq.(\ref{eq2}).

\begin{equation}
    \text{CV}_t^{(\text{DoF})}(\text{CV}_t^{(\text{Rank})}) = \text{DoF}_1^{(l)}(\text{Rank}_1^{(l)}) - \text{DoF}_t^{(l)}(\text{Rank}_t^{(l)})
\label{eq1}
\end{equation}

\begin{equation}
    \text{MCR}_t^{(\text{DoF})}(\text{MCR}_t^{(\text{Rank})}) = \frac{\text{DoF}_t^{(l)} (\text{Rank}_t^{(l)}) - \min_t \text{DoF}_t^{(l)} (\text{Rank}_t^{(l)})}{\min_t \text{DoF}_t^{(l)}(\text{Rank}_t^{(l)})}
\label{eq2}
\end{equation}

\begin{figure}[htp!]
    \centering
    \vspace{-2mm}
    \subfigure[CNN(MNIST)]{
		\includegraphics[width=0.45\textwidth]{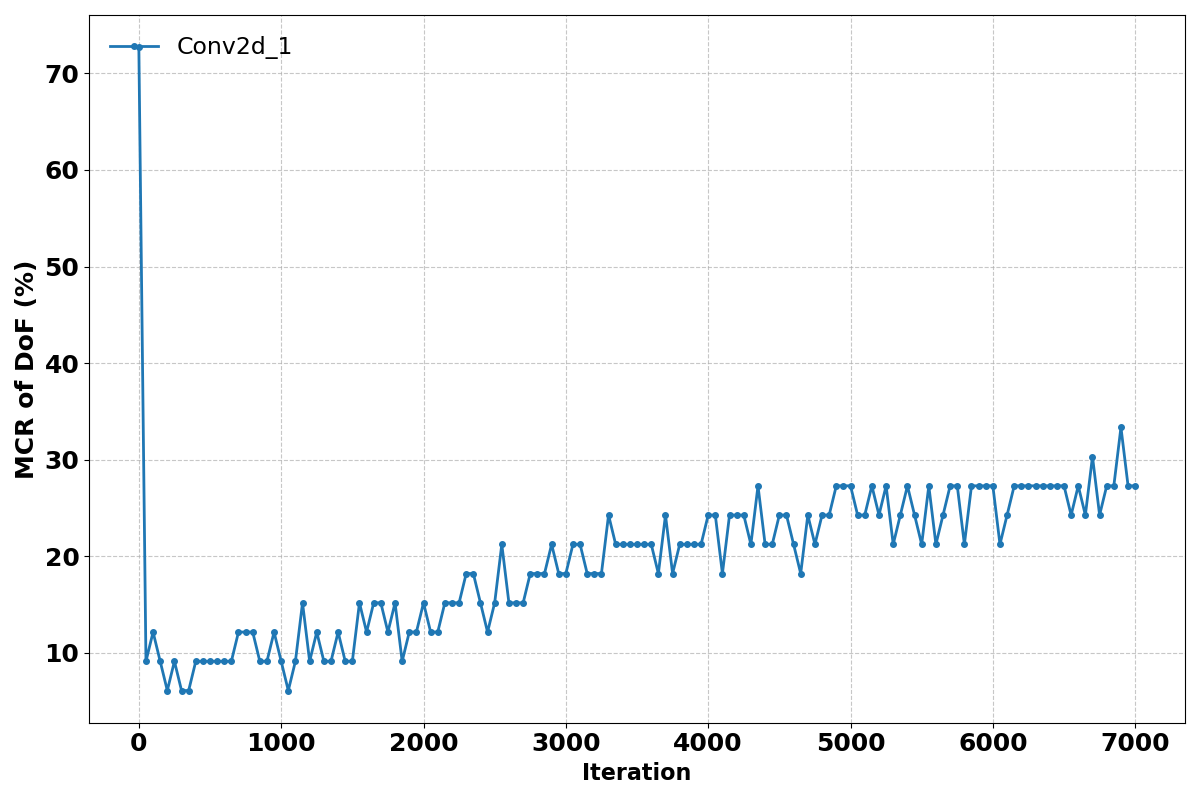}
		\label{DoFR(CNN)}
    }
    \subfigure[LeNet(CIFAR10)]{
		\includegraphics[width=0.45\textwidth]{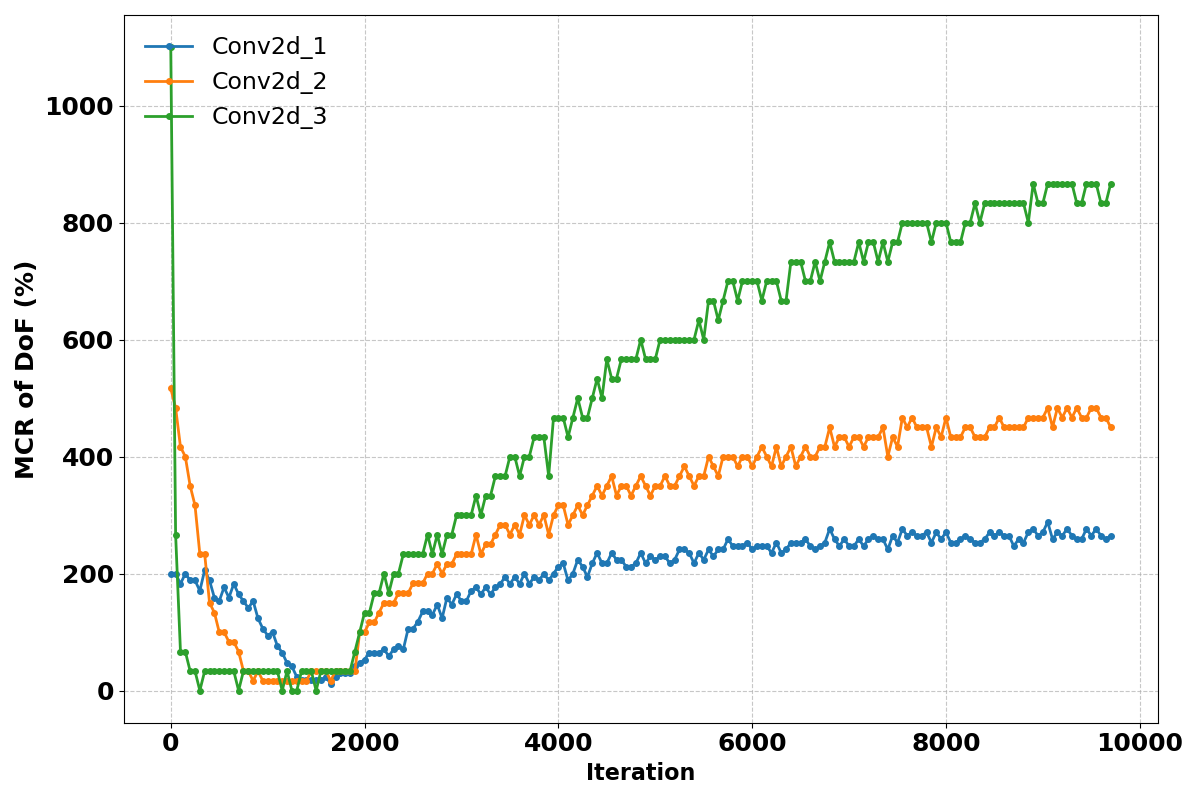}
		\label{DoFR(LeNet10)}
    }\\
    \subfigure[LeNet(CIFAR100)]{
		\includegraphics[width=0.45\textwidth]{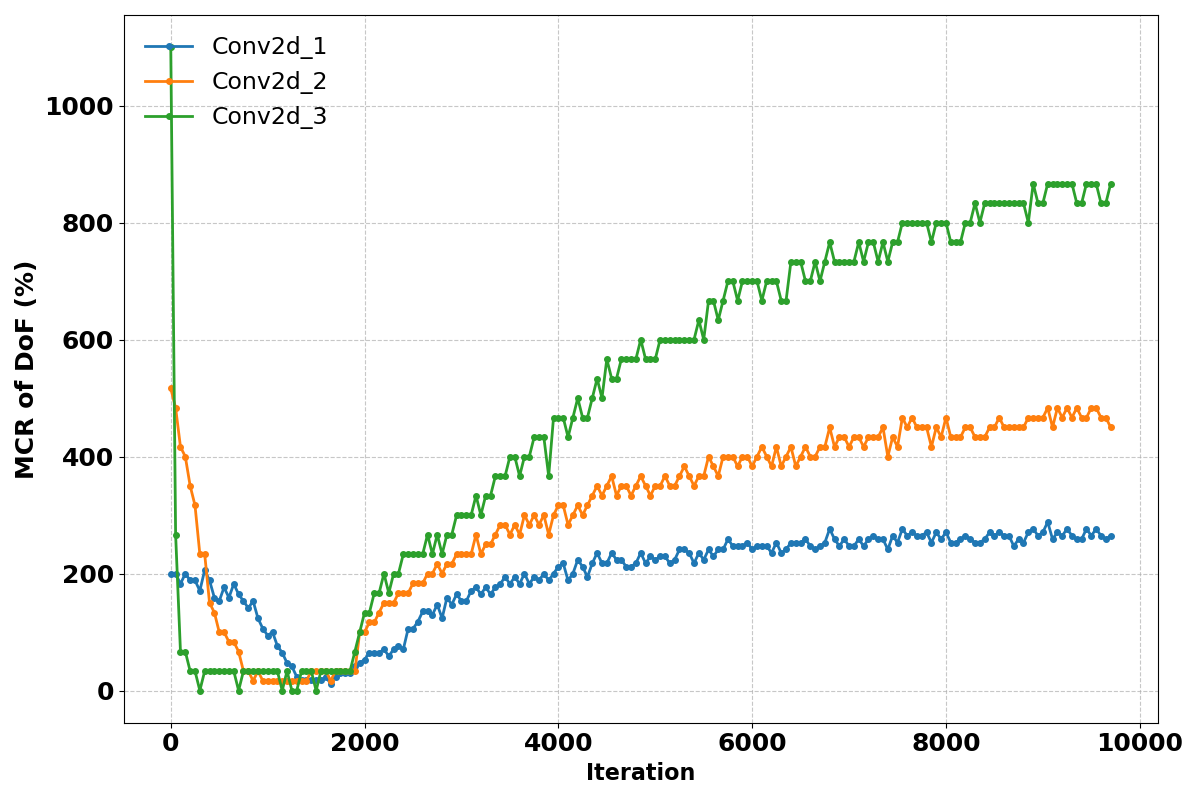}
		\label{DoFR(LeNet100)}
    }
    \subfigure[AlexNet(CIFAR100)]{
		\includegraphics[width=0.45\textwidth]{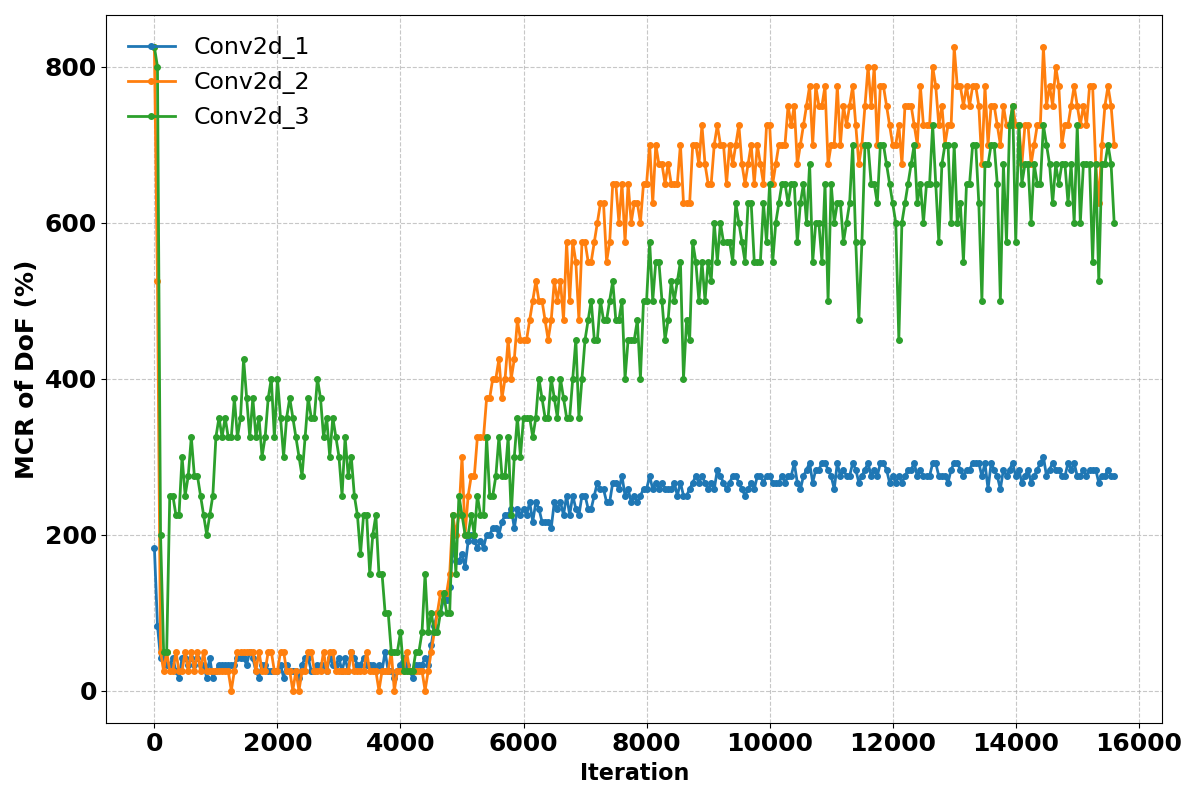}
		\label{DoFR(AlexNet1-3)}
    }
    \subfigure[AlexNet(CIFAR100)]{
		\includegraphics[width=0.45\textwidth]{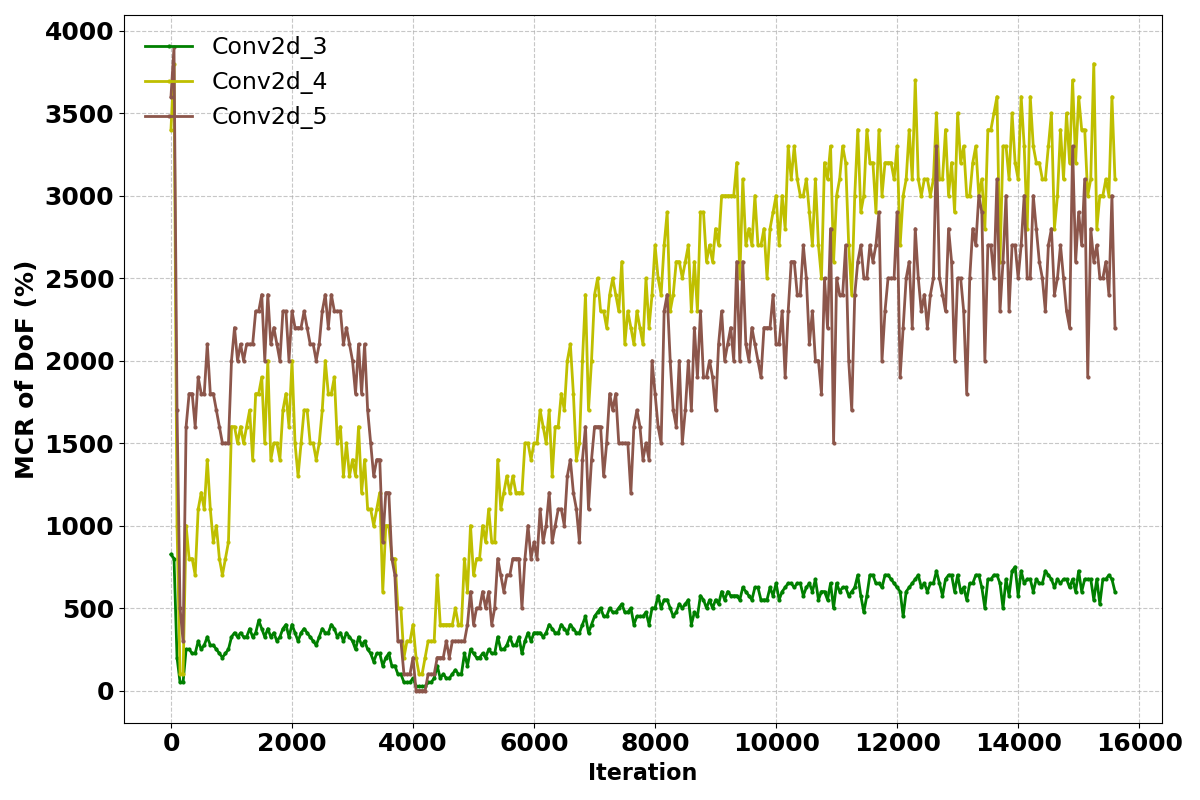}
		\label{DoFR(AlexNet3-5)}
    }
    \caption{MCR of DoF}
    \label{MCR of DoF}
\end{figure}

\begin{figure}[htp!]
    \centering
    \vspace{-2mm}
    \subfigure[CNN(MNIST)]{
		\includegraphics[width=0.45\textwidth]{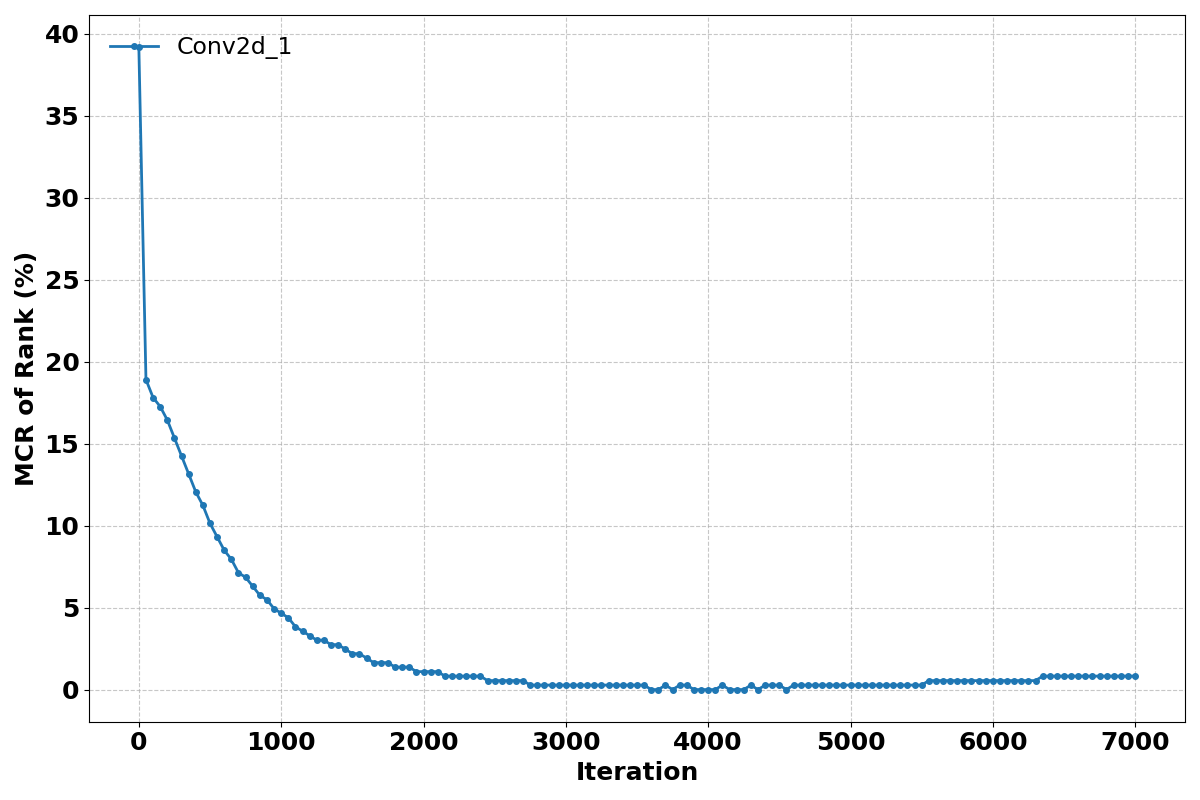}
		\label{RankR(CNN)}
    }
    \subfigure[LeNet(CIFAR10)]{
		\includegraphics[width=0.45\textwidth]{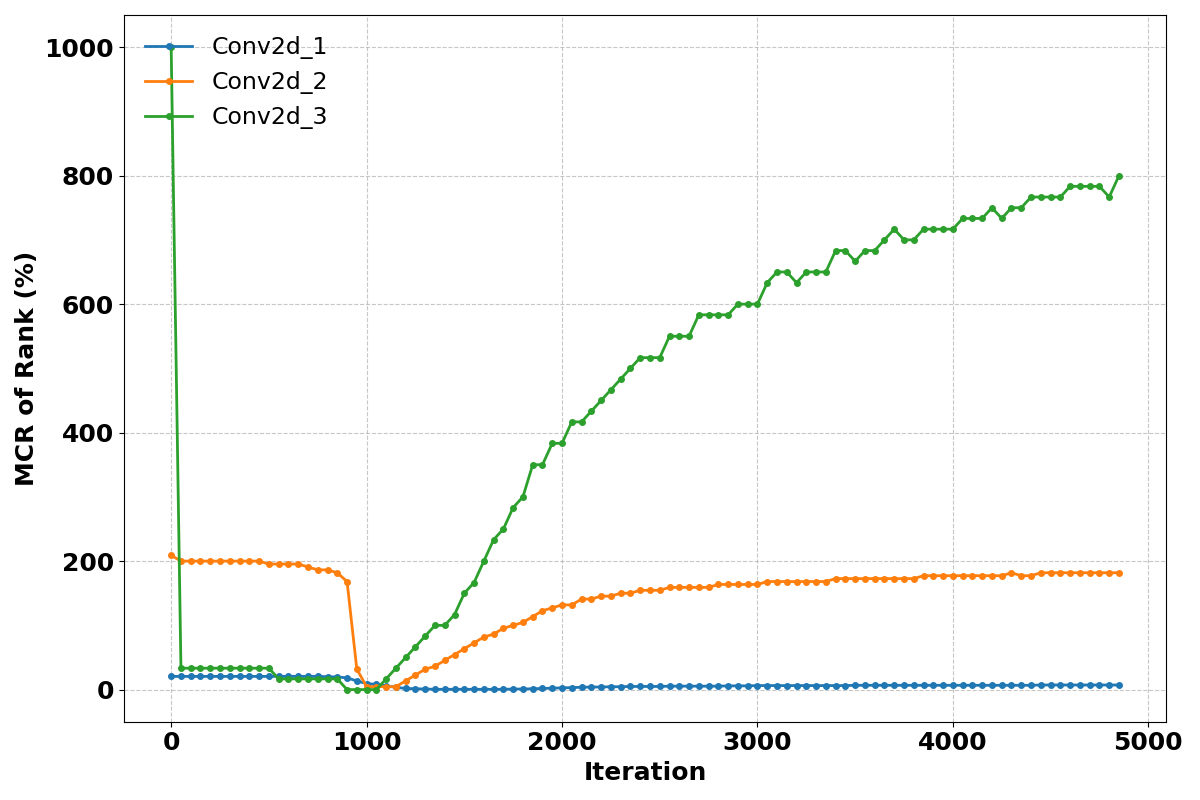}
		\label{RankR(LeNet10)}
    }\\
    \subfigure[LeNet(CIFAR100)]{
		\includegraphics[width=0.45\textwidth]{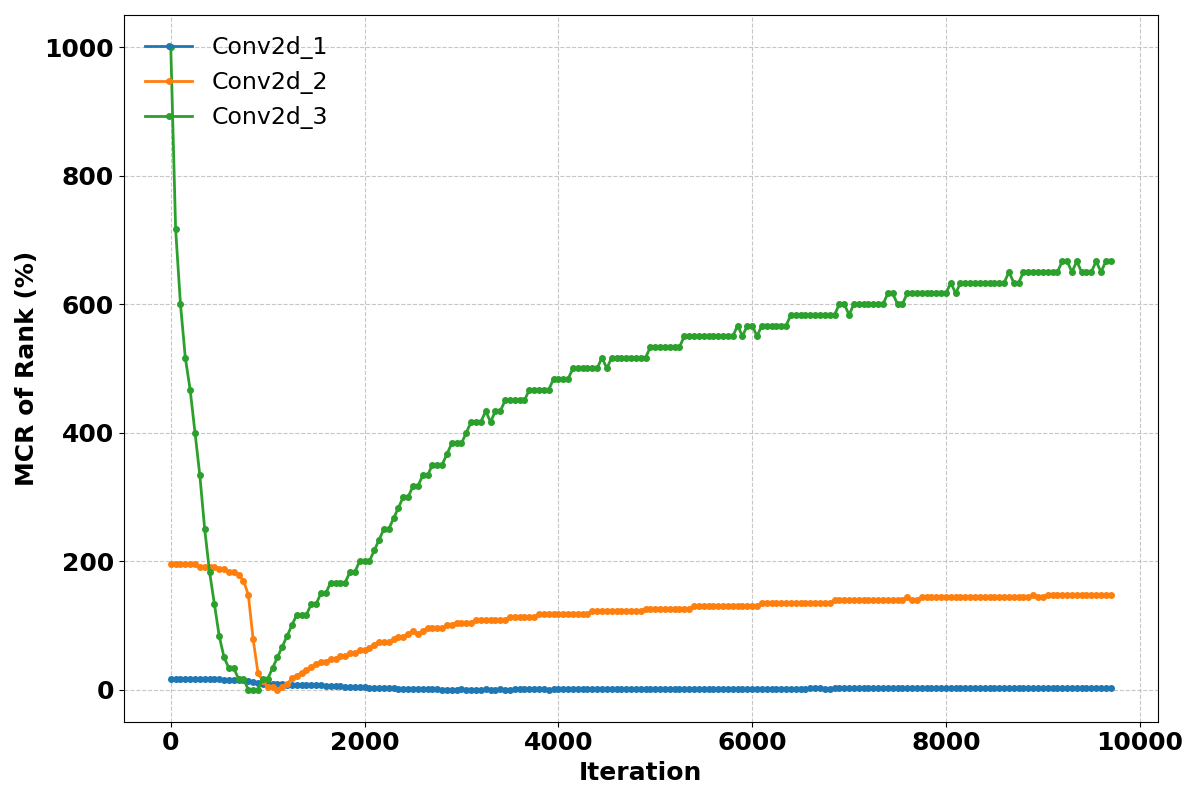}
		\label{RankR(LeNet100)}
    }
    \subfigure[AlexNet(CIFAR100)]{
		\includegraphics[width=0.45\textwidth]{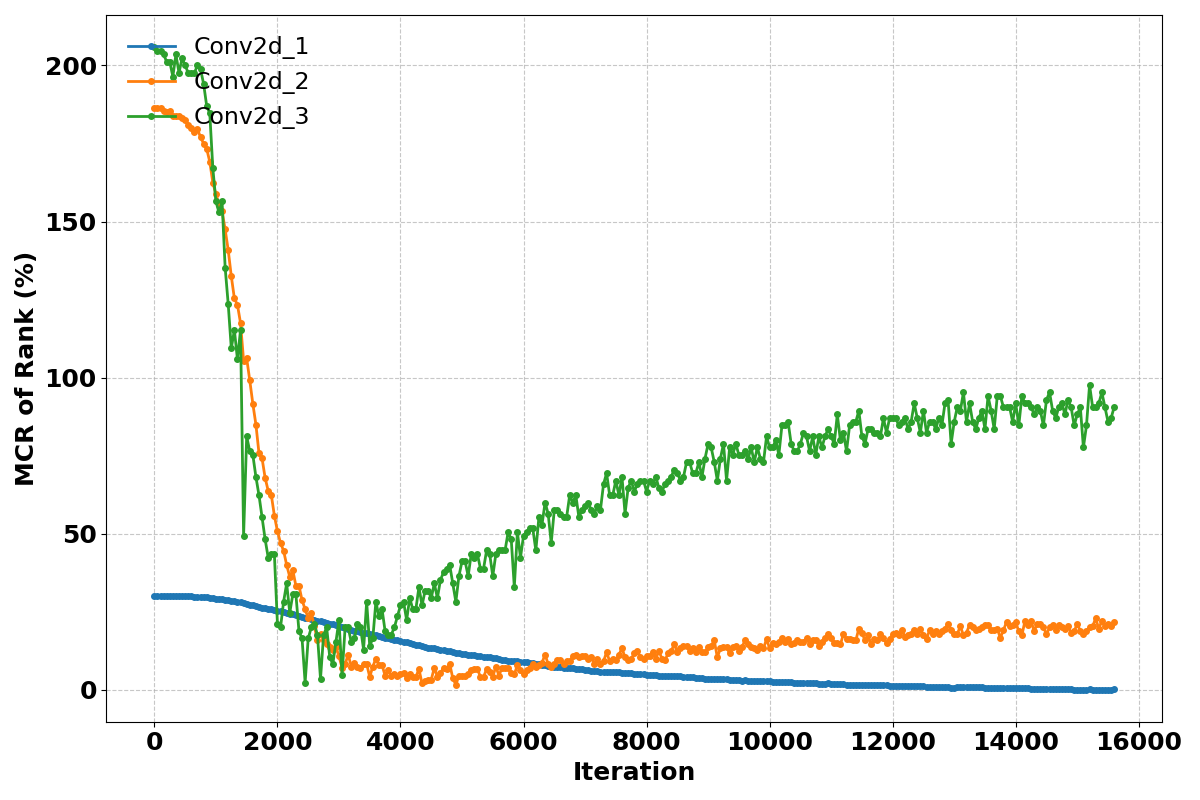}
		\label{RankR(AlexNet1-3)}
    }
    \subfigure[AlexNet(CIFAR100)]{
		\includegraphics[width=0.45\textwidth]{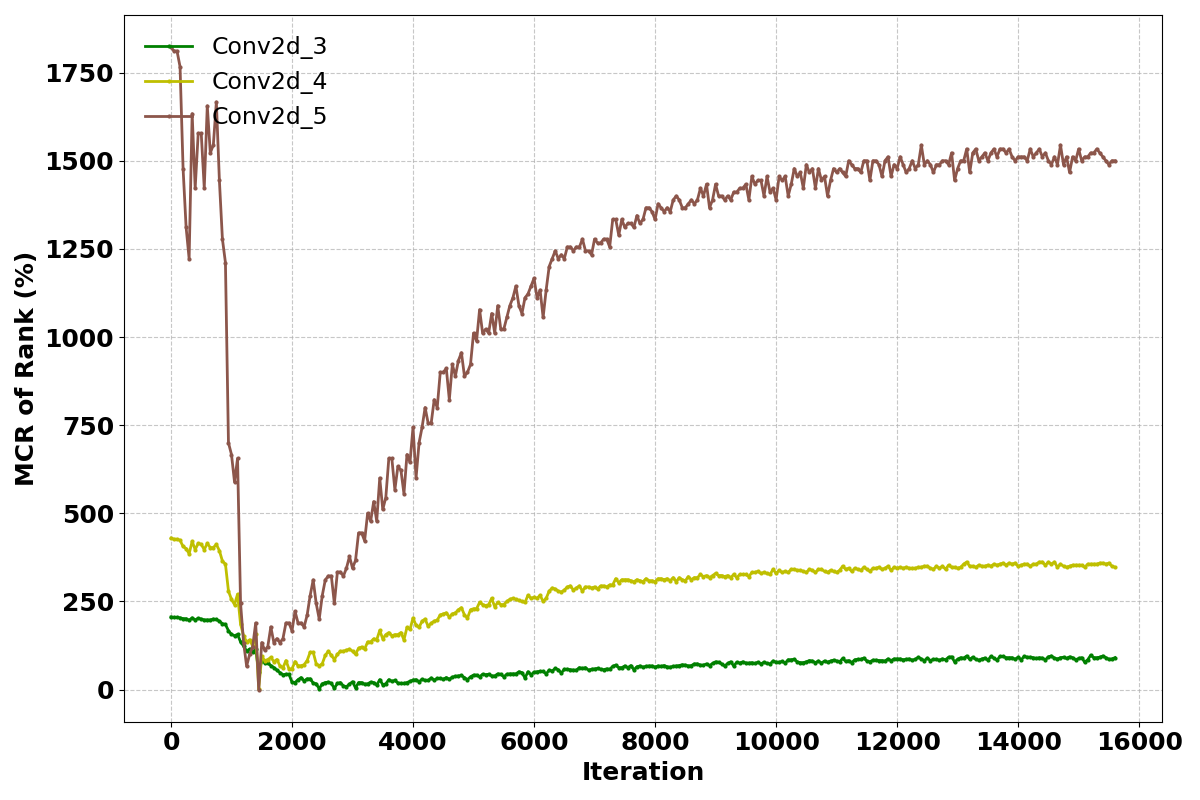}
		\label{RankR(AlexNet3-5)}
    }
    \caption{MCR of Rank}
    \label{MCR of Rank}
\end{figure}

CV represents the difference between the initial and current values of DoF or Rank, while MCR measures the relative change based on the minimum value during training. Fig. \ref{DoF} and \ref{Rank} depict the CV dynamics, with the maximum CV and its reduction relative to the final training stage summarized in TABLE \ref{table1}. Fig. \ref{MCR of DoF} and \ref{MCR of Rank} illustrate the MCR dynamics, and the final MCR values after training are shown in TABLE \ref{table2}. To better illustrate the observed patterns, we separately present the results of the first three feature extraction layers (Conv2d\_1, Conv2d\_2, and Conv2d\_3) and the last three feature extraction layers (Conv2d\_3, Conv2d\_4, and Conv2d\_5) of the AlexNet model.

The experimental results reveal consistent patterns. CV values for both DoF and Rank initially increase and then decrease as training progresses, reflecting a decrease followed by a recovery in DoF and Rank during this process. Layers closer to the output exhibit smaller reductions in DoF CV and larger reductions in Rank CV after reaching their peaks. These reductions are strongly correlated with the success rates of membership inference attacks, where layers with smaller reductions in DoF CV or larger reductions in Rank CV are more vulnerable. Additionally, MCR values generally increase during the later stages of training, with higher values observed in layers closer to the output. These layers also exhibit higher vulnerability to membership inference attacks, emphasizing the importance of monitoring MCR alongside CV in privacy risk assessments.

\textbf{Analysis.} The observed trends in DoF and Jacobian rank are indicative of the dynamic nature of information representation within deep learning models during training. The initial decrease in both metrics can be attributed to the model learning to extract and abstract general features from the input data, leading to a compression effect. This phase reduces the sensitivity of intermediate outputs to specific input variations, acting as an information bottleneck. Conversely, the subsequent increase in DoF and rank reflects the model's shift towards capturing more specific and detailed features of the input data, thereby expanding its information retention capabilities. This phase, marked by higher DoF and rank, signifies that intermediate outputs are more responsive to input variations and retain richer information, which could increase the risk of input data privacy leakage.

Our results suggest that both CV and MCR are effective indicators of privacy leakage risks. While Rank-based metrics provide more pronounced trends, they come with higher computational costs due to the need for additional differentiation steps. In contrast, DoF-based metrics are computationally efficient and suitable for scenarios where assessment accuracy is less critical. Smaller reductions in DoF CV and higher MCR values suggest greater privacy risks, while Rank-based metrics are more appropriate for scenarios requiring high-accuracy risk evaluations. These insights provide valuable guidance for developing privacy-preserving mechanisms and evaluating layer-specific privacy risks in deep learning models.

The relationship between the trends in DoF and Jacobian rank and input data privacy leakage is significant. During the initial phase of training, when the DoF and rank are lower, the intermediate outputs are less informative about the inputs, leading to a reduced risk of privacy leakage. However, as training progresses and these metrics increase beyond the baseline levels, the intermediate layers begin to carry more detailed information about the inputs. This heightened sensitivity and information retention pose a greater risk for privacy leakage, as adversaries may be able to exploit these outputs to reconstruct or infer input data. The use of baseline comparisons for DoF and rank helps in assessing how much additional input information is being retained over time, thereby identifying critical training periods when the model is more vulnerable to privacy risks.

\section{Conclusion and Future Work}
This study introduces a novel framework for assessing privacy risks in intermediate outputs of deep learning models by leveraging Degrees of Freedom (DoF) and the rank of the Jacobian matrix. Our analysis demonstrates that the dynamic changes in DoF and Jacobian rank during training reveal significant insights into the sensitivity and information retention of intermediate layers. Metrics such as CV and MCR effectively quantify these changes, highlighting critical layers with heightened vulnerability to privacy attacks. Experimental results indicate that layers with higher MCR values and specific CV trends are more susceptible to membership inference attacks, providing a reliable basis for evaluating privacy risks. These findings not only underscore the importance of monitoring intermediate outputs during training but also offer practical guidelines for designing privacy-preserving mechanisms.

While our framework provides an efficient alternative to computationally expensive attack simulations, several areas require further investigation. First, the scalability of our approach to more complex architectures and larger datasets warrants exploration. Extending the methodology to transformer-based models and federated learning settings could address diverse real-world applications. Second, while we relied on Gaussian projections and threshold-based eigenvalue selection for computational efficiency, optimizing these parameters for specific models and tasks could enhance accuracy. Additionally, integrating our metrics with existing privacy-preserving frameworks, such as differential privacy, may yield comprehensive solutions for mitigating privacy risks.

Future research should also explore the theoretical underpinnings of the observed relationships between DoF, Jacobian rank, and privacy vulnerability. Developing formal guarantees on the linkage between these metrics and attack success rates would strengthen the reliability of our framework. Lastly, incorporating adaptive mechanisms to dynamically adjust model training based on real-time privacy risk evaluations could pave the way for proactive privacy-aware training paradigms. These directions will ensure that the proposed framework remains robust, adaptable, and impactful in safeguarding data privacy within deep learning systems.

\bibliographystyle{named}
\bibliography{neurips_2024}

\end{document}